# Using Word Embeddings Based on Psychological Studies of Fairness to Develop a Fairness Metric for Texts

Ahmed Izzidien, David Stillwell

Fairness is a principal social value that can be observed in civilisations around the world. A manifestation of this is in social agreements, often described in texts, such as contracts. Yet, despite the prevalence of such, a fairness metric for texts describing a social act remains wanting. To address this, we take a step back to consider the problem based on first principals. Instead of using rules or templates, we utilise social psychology literature to determine the principal factors that humans use when making a fairness assessment. We then attempt to digitise these using word embeddings into a multi-dimensioned sentence level fairness perceptions vector to serve as an approximation for these fairness perceptions. The method leverages a pro-social bias within word embeddings, for which we obtain an F1= 81.0. A second approach, using PCA and ML based on the said fairness approximation vector produces an F1 score of 86.2. We detail improvements that can be made in the methodology to incorporate the projection of sentence embedding on to a subspace representation of fairness.

## Introduction

Given the centrality of texts in the digital humanities, recent work has attempted to leverage the use of word embeddings to reveal information which would otherwise not be readily available. The process, which we detail later on, is based on representing words based on their co-occurrences with other words (Mikolov et al. 2013; Rong 2014; Pennington, Socher, and Manning 2014). The representation can be thought of as a digital object that captures meaning (J. E. Dobson 2021). Recent work by (Kozlowski, Taddy, and Evans 2019), for example, employ word embeddings to separate between associations carrying cultural meaning across different time periods, i.e., *rich – poor*, *affluence – poverty*. Word embeddings have also been applied to official state inquiries on social justice (Leavy, Keane, and Pine 2019), as well as on texts related to the COVID-19 epidemic (Aiello et al. 2021). Likewise, (Jha, Liu, and Manela 2020) use them to measure popular sentiment towards finance across several decades in which they consider dimensions such as the *financial system hurts the economy- financial system helps the economy* to separate between attitudes. While natural language processing (NLP) for textual analysis has been used across a number of psychological domains such as personality detection (Youyou, Kosinski, and Stillwell 2015), a search through the literature for a metric that delivers a fairness score when it is used against sentences produces no records.

In this paper, we ask the question of whether or not it is possible for software to approximate typical fairness perceptions and incorporate them into a measurement tool for simple descriptions of social interactions, one which would allow a sentence describing an interaction between two or more individuals to be classed as *fair* or *unfair*?

Instead of using a philosophical template to define what is *fair* or *unfair*, or by specifying a particular kind of fairness to measure, we approach the problem based on first principles: What are the factors that humans typically use when making a fairness assessment, as found in controlled psychology studies? Is it possible to use these factors in vectors to act as measures of digital texts?

In doing so, we aim to approximate those perceptions, which together form a basis for a *fairness perception*, which we hypothesise will allow sentences to be classed according to which perception they are closer to, being *fair* or *unfair*.

Although the paper does not set out to produce a fully validated and verified fairness measurement tool for documents, it contributes to the development of one based on an approximation of these factors that humans engage when making such measurements. As such, we do not claim to be measuring a specific fairness type, e.g., distributional/outcome. However, as will be discussed, fairness evaluations engage a number of principal psychological factors, and it is these factors that we attempt to approximate using a method of word embeddings and vector arithmetic. While the ML techniques used in this paper are well established, our approach to digitising the factors, and the theory behind their use in this domain is new. Not least as no such measure exists in the literature.

On the closely associated topic of morality, a number of papers have investigated the use of Moral Foundation Theory (MFT) (Graham et al. 2013) to analyse texts. The general approach has been to label a dataset of texts with categories, and apply a ML algorithm to learn the distinctions between each category (Rezapour, Shah, and Diesner 2019; Hoover et al. 2020; Rezapour, Dinh, and Diesner 2021). Similarly, (Pennebaker, Francis, and Booth 2001; Araque, Gatti, and Kalimeri 2020) used pre-defined measures of moral language. These approaches have proved useful as a form of topic modelling of language, yet they are unable to mark a sentence as being fair or unfair, or offer a degree of explainability as two why a classification was made, and to what degree each sentence is fair or unfair. Indeed a common challenge to ML systems of this type is that of the limitations imposed by the technology on explainability (Danilevsky et al. 2020; J. Dobson 2020).

Such limitations also present when (Ruggeri et al. 2021; Lagioia et al. 2019) class documents as belonging to legally fair or unfair categories based on pre-trained legal clauses. They do not seek to identify explainable factors behind why something is *perceived* as fair or unfair, but instead, accept it as given that certain legal clauses are fair, for which they implement ML methods to draw a classification boundary. Testing their API ("CLAUDETTE" 2021) with the sentence '*the boy will hit the girl*', for example, produces the result: 'Claudette found no potentially unfair clause'. Which may be a reflection of out-of-domain knowledge limitations.

Work done by (Schramowski et al. 2019a) and (Jentzsch et al. 2019) have further demonstrated that language models (LM) hold implicit representations of moral values. They do so using vector comparisons based on a template of Do's and Don'ts. Furthermore, (Schramowski et al. 2019a) replicates the moral choices found by (Jentzsch et al. 2019), then computes the variance explained by another LM, the Universal Sentence Encoder (USE) (Cer et al. 2018) with respect to Yes-No question templates on moral choices. Further work in (Izzidien 2021) replicates the finding that word embeddings contain implicit biases, and proposes to use them to make assessments of verbs.

Building on these studies, we propose to harness these implicit moral social biases to act as a metric for an explainable assessment of sentences, specifically those related to fairness. The paper is organised as follows. The methods section is presented next, this incorporates a detailed study to determine the most explanatory psychological factors present in fairness assessments. The paper then details two approaches to digitise these psychological factors using word embeddings and ML. The results are then presented. A short discussion is followed by improvements, limitations, and the conclusion.

**Methods**

To be able to characterise the factors that humans use when making a fairness assessment, we turn to the psychology literature. Using controlled experiments, social psychologists have considered what factors best explain pro-social acts such as fairness. These studies have involved between-subjects trials and experimental variable manipulations as detailed next.

**The Principal Factor**

One of the challenges of exploring which traits positively predict people acting fairly, is that social interactions often involve feedback between individuals. Thus, asking a participant to share part of a resource with another introduces many confounds, such as social desirability (Platow 1994), and possible expectations of reciprocity (Fehr and Gächter 2000). To address this problem, psychologists attempted to isolate factors that predict prosocial acts by modelling a scenario, in which an actor has the choice to share without any concern for the repercussions of withholding. This has taken the form of the Dictator Game (DG) (Guala and Mittone 2010).

A DG allows a person to choose, how much of a resource to share with another person, without any concern of being punished, allowing for the removal of strategic intentions (Ibbotson 2014). A person is typically presented with a pot of cash, which they may keep in its entirety. They may also share part or all of the cash with another player, or several players. The context and frame of the study is often manipulated by the researcher to attempt to decipher, which factors influence how much is given. Spanning 25 years, and 20,813 trials, incorporating 24 factors to overcome the limitations of single studies, a meta study by (Engel 2011) was conducted on these games. They determined that outside of age the strongest positive effects for two person DGs concerned the two variables of recipient *need* and *legitimacy*.

Using the effect sizes (marginal effects) from the meta-analysis as the true population effect size, a second meta-analysis (Ortman and Zhang 2013) calculated the post-hoc statistical power for the studies included in Engel's meta-study -which investigate at least one of those explanatory variables by the non-central $t$ distribution. They found the effect size for *deserving recipient* was 1 for four of the studies and above 0.6 for the fifth. *Recipient earned* was under 0.2, *dictator earned* was close to 0.6 for eleven studies. Outside of a framing effect (Zhang and Ortmann 2014) also replicates (Engel 2011).

Given that these two factors, *legitimacy* and *need*, were two main psychological contributors to giving, many of the individual studies that found this effect were characterized by their use of the language of *need*, *deservedness* and *entitlement* (Cappelen et al. 2013). A study using the frame: "Note that he relies on you" found that selfish behaviour in the DG almost vanished (Brañas-Garza 2007). Further, a strong effect was observable that was independent of the extent of altruism measured or of the dictator being seen (Rodrigues, Ulrich, and Hewig 2015). The perception of fairness was demonstrated as being modulated by an integration of the two factors of *egalitarian motivation* and that of *entitlement* (Feng et al. 2013). On entitlement effects, acts of giving were found due to the sense of *earned* shares as evocative of a right that they *deserve* (Cappelen, Sørensen, and Tungodden 2010). Such entitlement frames have also been used to explain the observation that individuals in such contexts do not share more of their earned income with those in greater need (Eckel and Grossman 1996).

It appears that the language encompassing *need* and *entitlement* is evocative of two social values: a right e.g., *he worked for it*, and a responsibility to help: e.g., *he relies on you*, respectively. Both rights and responsibilities may be considered opposite sides to the same coin: If someone has the right to something, then someone else has a responsibility towards that person with respect to that right. As such, responsibility is considered concomitant to a right, as is well established legal philosophy (Kramer 2000).

Using 150 observations (Tisserand, Cochard, and Le Gallo 2015) analysed the two person DGs across seventy papers (1986 to 2014). Their comparative pooled meta-analysis revealed that dictators from countries low in industrialisation exhibited greater considerations for fairness. Industrialization had a strong negative and significant influence on share. Players from industrialized countries shared significantly less. This was confirmed in Engel (2011) who found that with indigenous countries, a proposer gives more. Such may be reflective of the characteristic of responsibility, which studies report to be influenced by the cultural climate of the person: In a 40 years longitudinal cohort, responsibility was at its lowest when a culture of individualism was at its peak (Helson, Jones, and Kwan 2002; Jensen-Campbell, Knack, and Rex-Lear 2009; Tisserand, Cochard, and Le Gallo 2015). Indeed a study (Handgraaf et al. 2008) that specifically manipulated the DG to account for mediation effects found that the trait of *social responsibility* was the best predictor of giving.

Given this principal factor of responsibility, we turned to the wider literature to consider studies that specifically controlled for *responsibility* in their manipulations. These studies were also found to replicate the above finding. A study by (van Dijk and Vermunt 2000) asked participants as to the extent they considered it was their responsibility to share the money fairly. They found the unilateral power distribution in the DG triggered a social responsibility norm. The paper found a main effect that those in the DG condition felt more strongly they ought to share their money fairly than did those taking part in the Ultimatum Game setting. A study by (Yang et al. 2020) found a positive correlation between a sense of community responsibility (SOC-R) and altruism behaviour (AB). Their regressions demonstrated a linear relationship between them, with SOC-R as the predictor and AB as outcome. In a study by (Brañas-Garza, Durán, and Paz Espinosa 2009), factors of personal involvement and responsibility explained the reasons behind why positive values were given in DGs.

(Sijing and Jianhong 2011) used a DG and Third-Party Game to activate the social norm of fairness. They found social responsibility had a critical role in norm activation. Players who scored higher on responsibility were characterized with greater pro-social behaviour after being activated. A study by Milgram determined that when one was able to make another person responsible for an act, anti-social acts could more easily materialise. Concordantly, (Cui et al. 2015) reports that the activations of a person to witnessing others in pain is modulated by the witnessing parties responsibility, whereby responsibility sharing, or not being responsible, lowers the pain-matrix neural activity.

One method to attempt to falsify the claim that responsibility plays such a central role, would be to remove or diminish it. A number of studies attempted this manipulation. These are detailed next.

A study by (Cryder and Loewenstein 2012) considered whether individuals were more generous in two player DGs than in conditions for which responsibility for any one receiver is potentially divided across more than one dictator. When an individual was completely responsible for somebody else's outcome, the chances of giving rose by a factor of 3.03 ($\chi^2$(df=1, N=80) =5.58, p < 0.02). Unambiguous responsibility for a single receiver leads to a higher share. Using a shopping area, a condition was set to elicit a sense of responsibility. Those solely responsible for the outcome of another individual were found to be significantly more generous. In (Hamman, Loewenstein, and Weber 2010), delegated agents led to settings in which the accountability for questionable moral decisions become diffused, whereby no single person was seen as responsible. Dictators generally preferred to delegate, which led to highly reduced amounts being shared with others.

The indirect assessment of responsibility was made by (Bartling and Fischbacher 2012). Using a 'punishment assignment' for the results of decisions, the measure of responsibility outperformed measures that used inequity aversion or reciprocity to predict punishment behaviour. Lastly, a study (Charness 1998) found that participants respond with more generosity when a wage is determined by a random process than when assigned by a third party. Such a shift in perceived responsibility for the pay was found to alter behaviour. Participants felt less of an impulse to contribute ($\chi^2$ = 20.6, df = 9) to an anonymous employer when they perceived that a third party had approved the wage in some way resulting in a shift of some responsibility for the determination of the outcome. They found that individuals are in general more generous with anonymous strangers when they must assume full responsibility for payoff allocation.

**Contingent Factors**

When a human perceives a context as one that warrants a *responsibility* evaluation (Handgraaf et al. 2008), such evaluation is dependent on contingent factors. By contingent factors, we mean the principal factors needed to allow for a perception of *responsibility* to materialise. Intuitively, these are a perception of the frame (Engel 2011; Zhang and Ortmann 2014) in terms of:

1) The benefit-harm gained: A measure of how the actors actions will result in a benefit to the receiver or lack thereof (Brañas-Garza et al. 2014; Perera, Canic, and Ludvig 2016; Bruner and Kopec 2018; Chiaravutthi 2019).
2) The consideration of wider public benefit and harm (Sigmund, Hauert, and Nowak 2001; Gillet, Schram, and Sonnemans 2009; Lejano and Ingram 2012).
3) The emotional salience of the context: how much joy-pain is involved (Batson et al. 1991; Scheres and Sanfey 2006; Tabibnia and Lieberman 2007; Edele, Dziobek, and Keller 2013).

4) Outside the DG, a further perception of the possible consequences is incorporated: rewards and punishments (Nesse 1990; El Mouden et al. 2012, 24; Boyd et al. 2003; Scheres and Sanfey 2006; Henrich et al. 2001; Bartling and Fischbacher 2012; Strang and Park 2017).

These principal factors interact in a social context, allowing for a pro-social human propensity, or pro-social bias, to materialise, termed the ultra-cooperative trait, seen as unique to human society (Nowak 2006) (Tomasello 2014). It is these factors that we will use in word embeddings to act as measures.

**Language and Pro-Social Factors**

The use of language has been shown to reflect social perspectives (Kennedy et al. 2021). It has also been shown that a variety of social biases found in the usage of language can be measured when they are used in word embeddings owing to co-occurrences (Pennington, Socher, and Manning 2014), such as demographic features (Kozlowski, Taddy, and Evans 2019) and ethic and gender biases (Garg et al. 2018).

Given a human propensity for pro-social actions, and its articulation in general discourse, such a pro-social propensity may also present in word embeddings, where certain types of social interactions are associated with praiseworthy terms, while others are associated with blameworthy terms, such as *fair* and *unfair* acts respectively. We detail this next.

**Word Embeddings as Measures**

**Approach 1**

One of the most pertinent features of word embeddings are their mathematical properties (Pennington, Socher, and Manning 2014). These words become represented by vectors (Mikolov et al. 2013; Pennington, Socher, and Manning 2014). This process of vectorization incorporates the quantification of word frequencies, probability values, and co-occurrence relations, among other possible options (J. E. Dobson 2021). Such vectors can also be added, subtracted and compared. Vectors can also be compared to each other using cosine similarity. Closely associated vectors score closer to +1, with less similar scoring closer to -1, allowing for a measure of how similar vectorised sentences are (Cer et al. 2018; Kozlowski, Taddy, and Evans 2019).

Given that language reflects the social values of its speakers (Smith 2010; Kennedy et al. 2021), we hypothesize that word embeddings will reflect the social propensities determined by the literature. Thus, sentences that describe fair acts will be more closely associated with sentences that describe responsibility, benefit, joy, and reward, than that of their antithesis terms of irresponsibility, harm, sadness, and punishment. Based on this, it becomes potentially possible to use this feature, this pro-social bias, as a metric. Actions that are typically hurtful will co-occur more with negative social evaluations in typical corpora, reflecting the human propensity towards pro-social acts. As such it may become possible to leverage this bias as a metric.

To use embeddings for this purpose, we propose the method of adding and subtracting vectors (Foley and Kalita 2016) for the purpose of narrowing the implicit ontological associations of the resulting vector. In using word embeddings, built without any explicit ontological labels, the vector representation of the corpus implicitly reflect ontological knowledge (Erk 2012; Bhatia 2017; Runck et al. 2019; Racharak 2021). For example, grammatical ontologies become reflected due to the co-occurrence of specific grammatical knowledge in the co-occurrence of words (Qian, Qiu, and Huang 2016). The term *fairness*, being a collection of several social ontologies may be nominally represented using vectors, through the addition and subtraction of vectors which represent those factors found in the above psychology literature. We use this assumption to 'triangulate' a term, i.e., *fairness*, by outlying its main ontologies. Thus, similar to how a Venn diagram intersects at a mid-point, we add and subtract the vectors that represent sentences carrying a nominal meaning of the principal factors. In effect, we attempt to incorporate latent vector representations resulting from such addition and subtraction. We detail our method next.

**The Vectors**

To represent the psychological factors as vectors, we constructed the following sentences that describe them (table 1), which we then converted into vector format using the USE (Cer et al. 2018). Notation wise, a sentence is represented with a lower-case letter, and its vector space embedding by that letter with an arrow on top. For instance, the sentence v = "*it was irresponsible*", its vector space embedding will be $\vec{v}$. In cases where no letter is assigned to a sentence, the vector embedding of a sentence is designated by placing an arrow on top of the sentence. For instance, $\overrightarrow{it\ was\ very\ irresponsible}$.

The wording of the sentences were induced from each of the above numbered lists under the *Contingent Factors*. Thus, the two opposite terms of benefit-harm (Brañas-Garza et al. 2014; Perera, Canic, and Ludvig 2016; Bruner and Kopec 2018; Chiaravutthi 2019) were constructed into: $\overrightarrow{"it\ was\ beneficial"} - \overrightarrow{"it\ was\ harmful"}$. In considering the wider public benefit-harm (Sigmund, Hauert, and Nowak 2001; Gillet, Schram, and Sonnemans 2009; Lejano and Ingram 2012), we constructed: $\overrightarrow{"it\ was\ beneficial\ to\ society"} - \overrightarrow{"it\ was\ not\ beneficial\ to\ society"}$. For the emotional salience of the context, i.e., how much joy-pain is involved (Batson et al. 1991; Scheres and Sanfey 2006; Tabibnia and Lieberman 2007; Edele, Dziobek, and Keller 2013), the sentence constructed was $\overrightarrow{"it\ was\ joyous"} - \overrightarrow{"it\ was\ sad"}$. Given that outside of a DG, the factors of reward and punishment are contingent factors (Nesse 1990; El Mouden et al. 2012, 24; Boyd et al. 2003; Scheres and Sanfey 2006; Henrich et al. 2001; Bartling and Fischbacher 2012; Strang and Park 2017), the following sentences were used: $\overrightarrow{"was\ free\ to\ and\ rewarded"} - \overrightarrow{"was\ sent\ to\ prison\ and\ punished"}$. As the word 'free' can also mean 'no charge', we used two opposite terms on each side of the scale to reflect both the material and abstract nature of the consequence, i.e., prison *vs.* being free (material), and punished *vs.* rewarded (abstract). Lastly, the principal factor found was framed: $\overrightarrow{"it\ was\ very\ responsible"} - \overrightarrow{"it\ was\ very\ irresponsible"}$ and given that the quality of 'responsibility' was the most pertinent explanatory factor in the psychology studies above – under *the Principal Factor*, this explanatory factor was used with the term 'very' to emphasise the range.

Other words which also carry similar meaning may also have been used - as similar words occupy similar regions in vector space (Erk 2012). How the wordings affect outcome is given in the section on limitations.

The wordings used are given in table 1.

| Factor | Wording for scale |
|---|---|
| Responsibility dimension | it was very responsible - it was very irresponsible |
| Emotional dimension | it was joyous - it was sad |
| Public benefit dimension | it was beneficial to society - it was not beneficial to society |
| Personal benefit dimension | it was beneficial - it was harmful |
| Consequence dimension | was free to and rewarded - was sent to prison and punished |

**Table 1.** Using the principal and contingent factors for vector wordings

The vectors were constructed:

$$\vec{v}^{(1)} = \overrightarrow{\text{"it was very responsible"}} - \overrightarrow{\text{"it was very irresponsible"}}$$

$$\vec{v}^{(2)} = \overrightarrow{\text{"it was joyous"}} - \overrightarrow{\text{"it was sad"}}$$

$$\vec{v}^{(3)} = \overrightarrow{\text{"it was beneficial to society"}} - \overrightarrow{\text{"it was not beneficial to society"}}$$

$$\vec{v}^{(4)} = \overrightarrow{\text{"was free to and rewarded"}} - \overrightarrow{\text{"was sent to prison and punished"}}$$

$$\vec{v}^{(5)} = \overrightarrow{\text{"it was beneficial"}} - \overrightarrow{\text{"it was harmful"}}$$

The linear arithmetic allows for a capture of a range, going from positive to negative. Thus, if we consider the vector describing $\overrightarrow{\text{"it was beneficial"}} - \overrightarrow{\text{"it was harmful"}}$, and compare it to a vectorised test sentence, such as $\overrightarrow{\text{"the guard helped the man"}}$ through a cosine similarity calculation, the result will be a score from +1 to -1. The more associated the sentence is with benefit, the closer to 1 will be the result. Whereas sentences that are more associated with harmfulness will provide an outcome closer to -1

The sentence level *fairness perception* vector $\vec{v}$ is made by combining the vectors above:

$$\vec{v} = \vec{v}^{(1)} + \vec{v}^{(2)} + \vec{v}^{(3)} + \vec{v}^{(4)} + \vec{v}^{(5)}$$

We refer to this as the fairness vector, notwithstanding the limitations described earlier. In using this result, it becomes possible to compare $\vec{v}$ to the embedding of a test sentence, e.g., "the boy hit the baby" to determine how close the test sentence is in vector space to the parsimonious representations of fairness, by computing the cosine similarity.

In performing the linear manipulation – the addition and subtraction of vectors, the new vector $\vec{v}$ is able to capture a scale. One that allows for a comparison of a combination of these social dimensions to the sentence being tested.

Were it the case that only one of these social dimensions be used with a test sentence, the result would expectantly not capture the minimum pertinent factors associated with a perception of fairness. To consider this, the results of using each factor $\vec{v}^{(1)}$ to $\vec{v}^{(5)}$ separately are plot in the results section for comparison. The vectors were used against a list of 200 sentences compiled by three independent contributors.

A test is also conducted to compare the result of using such a parsimonious representation of fairness $\vec{v}$, against the result obtainable when using a straightforward vector embedding $\vec{v}^f$ instead:

$$\vec{v}^f = \overrightarrow{\text{"it was fair"}} - \overrightarrow{\text{"it was unfair"}}$$

It may be that such a vector ($\vec{v}^f$) will reflect variations on how the term 'fair' and 'unfair' is used in a corpus. Given the variation of definitions, it would be expected that such a representation would produce conflicting results. This contrasts with building up an ontology of fairness using representations commonly exhibited by humans as determined in the literature review above.

To consider whether the use of the fairness vector $\vec{v}$ is simply replicating a sentiment analyser, we perform a sentiment analysis using Vader (Hutto, 2021) and correlate the result with the result of using the fairness vector $\vec{v}$.

**Approach 2**

While adding and subtracting vectors offers a potential method to encompass fairness perceptions into a single vector, some information is inevitably lost by such a reduction. As an alternative we preserve the vectors for each of the evaluations, each as separate dimensions.

As such, we do not perform the above addition of $\vec{v}^{(1)} + \vec{v}^{(2)} + \vec{v}^{(3)} + \vec{v}^{(4)} + \vec{v}^{(5)}$, but rather use each independently. Thus, to evaluate a test sentence, e.g., "the shopkeeper assisted the customer", its word embedding vector $\vec{s}$ is compared, through cosine similarity, with the each of the five vectors $\vec{v}^{(1)}$ to $\vec{v}^{(5)}$, the results of which are stored in a multi-dimensioned vector $\vec{v}^m$. For example, supposing the result of such a cosine similarity operation were:

$(\vec{v}^{(1)}, \vec{s}) = 0.2$
$(\vec{v}^{(2)}, \vec{s}) = 0.1$
$(\vec{v}^{(3)}, \vec{s}) = 0.6$
$(\vec{v}^{(4)}, \vec{s}) = 0.3$
$(\vec{v}^{(5)}, \vec{s}) = 0.2$

The stored result $\vec{v}^m = [0.2, 0.1, 0.6, 0.3, 0.2]$

This is repeated for all test sentences, resulting in a dataset D1, which is then hand labelled with the correct fairness assessment (table 2):

| Index | Test sentence | Result $\vec{v}^m$ | Label |
|---|---|---|---|
| 1 | *the shopkeeper assisted the customer* | [0.2,0.1,0.6,0.3,0.2] | Fair |
| … | | | … |
| 200 | *the prisoner murdered the inmate* | $[-0.4, -0.6, -0.3, -0.3, -0.4]$ | Unfair |

**Table. 2** Snippet of dataset D1.

This produces a dataset containing the multidimensional vector and its label.

To explore how the factors induced from the psychology literature explain the data, a principal component analysis (PCA) is performed, initially with two components, then three.

To use the Dataset D1 for training a classifier, we perform ML using a logistic regression classifier, and a 1:7 test split.

To encode the sentences, we used the USE (Cer et al. 2018), detailed next.

**The Universal Sentence Encoder**

Initially shallow pre-training of early model layers became standard in NLP research through methods such as Word2vec (Mikolov et al. 2013). Subsequent progress followed trends similar to those in computer vision, which naturally led to pre-training of multiple layers of abstraction. These advancements resulted in progressively deeper hierarchical language representations, such as those derived using self-attention mechanisms in transformer-based architectures (Vaswani et al. 2017). Currently state-of-the-art NLP systems use representations derived from pre-training of entire language models on large quantities of raw text, and often involve billions of parameters. The success of neural network-based ML models, especially those involving very deep architectures, can be attributed to their ability to derive informative embeddings of raw data into submanifolds of real vector spaces. The common idea behind these developments is that we can learn syntax and semantics of natural languages by training a Deep Learning (DL) model in a self-supervised fashion on a corpus of raw text. Modern embedding methods combine word and sub-word (e.g., morpheme or character) level embeddings in a hierarchical and contextualized fashion to produce sentence and document level representations into (usually high-dimensional) submanifolds of $R^n$.

Given the high costs and low availability of manually labelled texts for training NLP models, word transfer models deploy pre-trained word embeddings (Mikolov et al. 2013; Pennington, Socher, and Manning 2014), which were successfully adapted to sentence-level representations (Conneau et al. 2017), and in particular utilised within the encoding module of the USE.

The USE introduces two embedding modules, one based on transformer sentence encoding providing high accuracy, and one deploying a deep averaging network (DAN), which focuses on computational efficiency.

This combination of features presented by the USE made it a good choice for our work. The reasons for this choice are two-fold. First, the mixture of deep self-attention induced encoding with the simpler DAN module, make it somewhat a compromise between predictive power and computational efficiency. Second, it marks a midpoint between sparser and more explainable models and deeper black-box architectures. This trade between explainability and accuracy is especially useful in the context of our work. Shallow models are closer to typical statistical learning and analysis procedures, which are prevalent in psychology and social sciences today, which make them ideal to study the ramifications of defining model components based on psychological theory.

Since the transformer side of the USE allows us to derive powerful context sensitive representations for natural language inputs, while on the other hand, the DAN side of USE allows us to inject these ethical considerations into the final representations of sentences produced by the combined encoder modules, it is particularly useful for work combining theory driven ethical considerations with natural language modelling methods. Our choice of USE allows us to impose knowledge derived from psychological findings. Such would be hard to do in a fully unsupervised setting. This has the further benefit of combining transparency and efficiency.

On a technical level, the USE first transforms languages to lower-case and tokenizes them via the PennTreebank (PTB) (Taylor, Marcus, and Santorini 2003). In both variants, a 512-dimensional embedding is produced. The transformer encoder deploys sub-graph encoding (Vaswani et al. 2017) to create sentence embeddings through a six-layered stack, whereby at each layer, a self-attention mechanism is followed by a feed-forward network. Words are fed through these layers, and their order as well as their context is taken into account through the use of positional embedding and sentence level attention mechanism. This process iteratively enriches representation of each word in order to augment the resulting embedding with contextual information of the sentence in which it appears within the corpus.

Each embedding is then added together, whereby the length difference of sentences is 'standardised' by dividing through the square root of the length. This results in an output sentence embedding in shape of a 512-dimensional vector, which is then fed into downstream tasks. The DAN variant is based on deep averaging networks (Iyyer et al. 2015) and follows a simpler approach, which starts by averaging embeddings for both bi-grams and words, and then passing these through a four-layered a neural network output module.

To ensure general purpose deploy-ability, the transformer encoding deploys multi-task learning, whereby one input model is fed into several downstream tasks. First, unsupervised learning is achieved through a Skip-Thought resembling task, replacing the encoder by the above two variants of input models (Kiros et al. 2015). Second, input-response task for parsed conversational data, which deploys the same encoder for input and output to model the difference of both, whereby their dot product determines the respective relevance, and it is fed through a softmax function, resulting in an optimisation over log likelihood of obtaining the correct response (Henderson 2017).

Last is the classification task using sentence pairs that represented the premises, hypotheses, and judgements about each pair. Such are fed through the transformer and DAN encoders described above, resulting in two 512-dimensional embeddings, processed by fully connected layers and three-way softmax, resulting in the probability of a judgement for each pair, which resembles earlier approaches (Conneau et al. 2017) to the task of natural language inference.

$$sim(\vec{u}.\vec{v}) = \left(\frac{1-arccos\left(\frac{\vec{u}.\vec{v}}{\|\vec{u}\|\|\vec{v}\|}\right)}{\pi}\right) \quad \text{Equ. 1.}$$

Finally, for classification transfer tasks, the respective outputs are fed into a specific deep neural network, whereas for the pairwise similarity task, the similarity is calculated in the following way:

First, the cosine similarity of two sentence embeddings coming from the two encoders is computed, then, the angular distance is obtained by applying the *arccos* function (Equ. 1) to the normalized inner product of the corresponding sentence representations.

**Results**

**Approach 1:**

We begin by considering 36 sentences (Appendix 1) selected randomly for graphical illustration purposes from the list of 200 sentences, each test sentence is compared through a cosine similarity with the vector:

$$\vec{v}^f = \overrightarrow{it\ was\ fair} - \overrightarrow{it\ was\ unfair}$$

This produces an incorrect result. Thirteen of the sentences are misclassed (figure 1). Correctly classed unfair sentences are classed in a manner that does not necessarily reflect typical evaluations: 'The man sickened the lady' is classed as closer to *'it was unfair'* than *'the father murdered the boy'* by orders of magnitude.

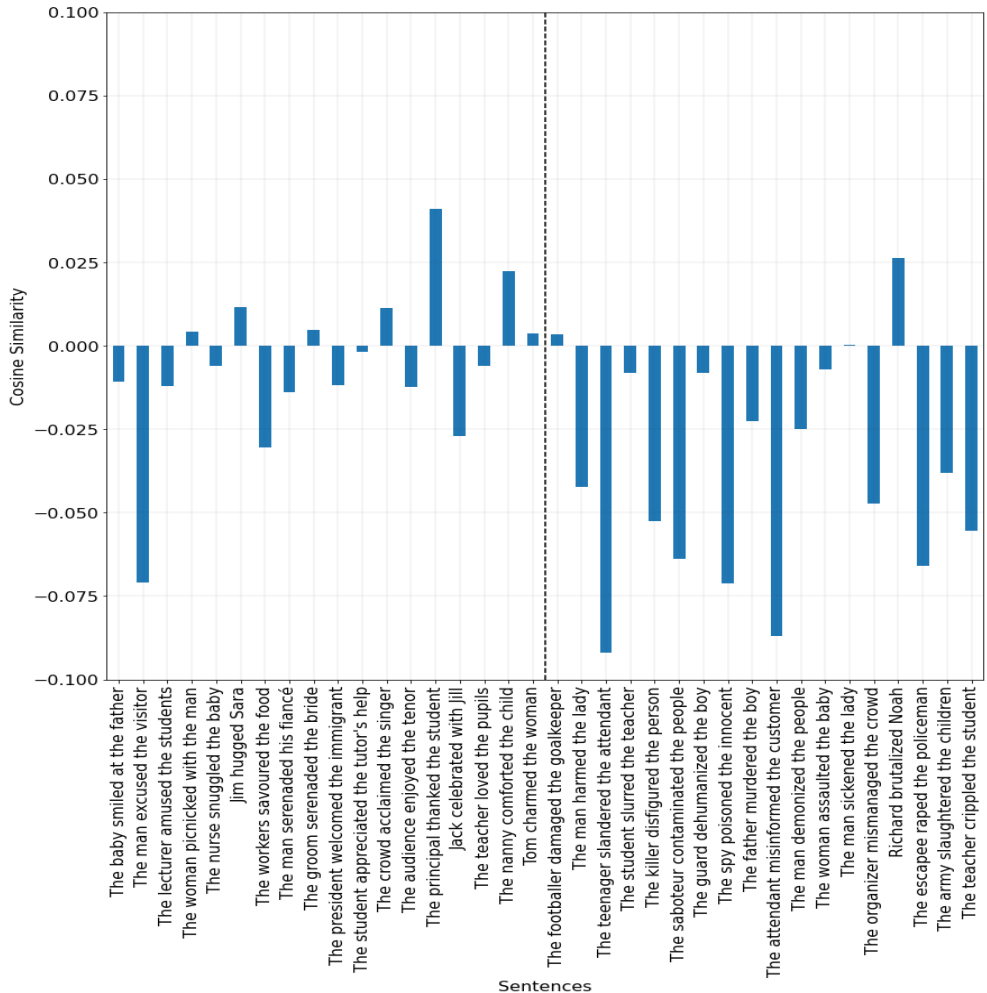

**Fig.1** Incorrectly classed sentences. All sentences of the left of the dotted line ought to be positive, while all sentences on the right ought to be negative. The incongruence of the scoring of the unfair sentences on the right can also be seen by comparing the score for *murder (-0.024)* to that of the act of *misinforming (-0.087)*.

If $\vec{v}$ is used instead, the results are shown in (figure 2).

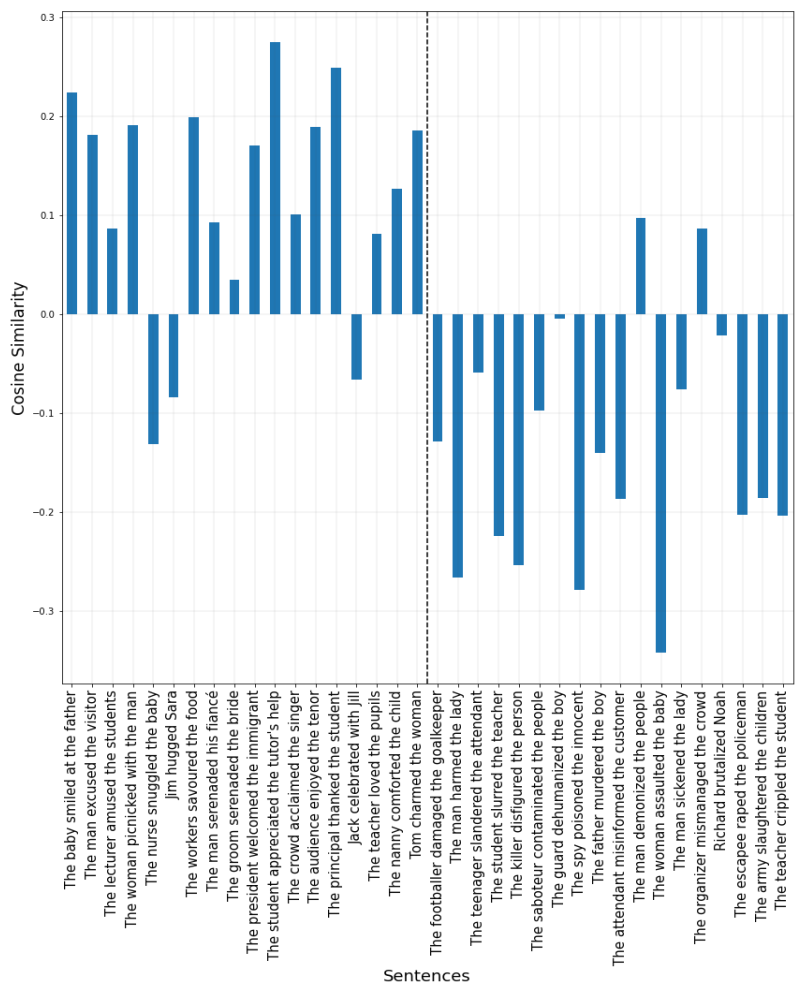

**Fig 2.** Use of the fairness vector $\vec{v}$ to measure the similarity of each sentence with a parsimonious representation of fairness.

The results (figure 2) may be said to be closer representation of typical fairness evaluations when compared to the results of figure 1.

As a means to demonstrate how each of the ontologies gives separate results when they are used signally, figure 3 is plot. This represents the outcome of using each of the five vectors: $\vec{v}^{(1)} to\ \vec{v}^{(5)}$ independently. Thus, for example, figure 3a reflects how similar each sentence is with the phrase: *"it was very responsible" - "it was very irresponsible"*. Six unfair sentences are misclassed as responsible. Each cosine similarity outcome is plot for each vector range as given in figure 3a to figure 3e. Here we observe how each sentence score is a reflection of factors contained within the corpus. For example, most of the test sentences describing fair acts, are classed as having a negative consequence (figure 3c).

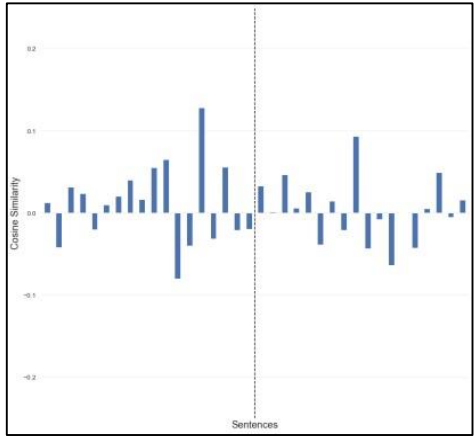

Fig 3a. Responsibility Dimension

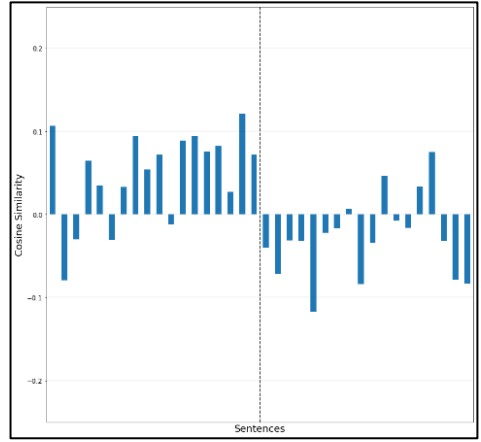

Fig 3b. Emotion Dimension

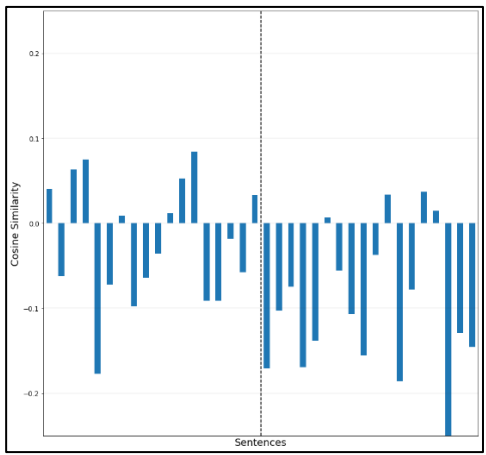

Fig 3c. Consequence Dimension

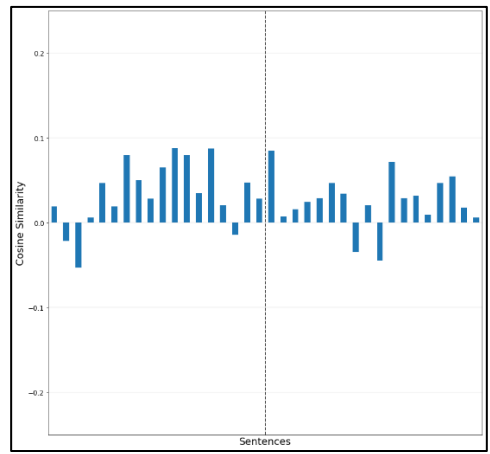

Fig 3d. Benefit Dimension

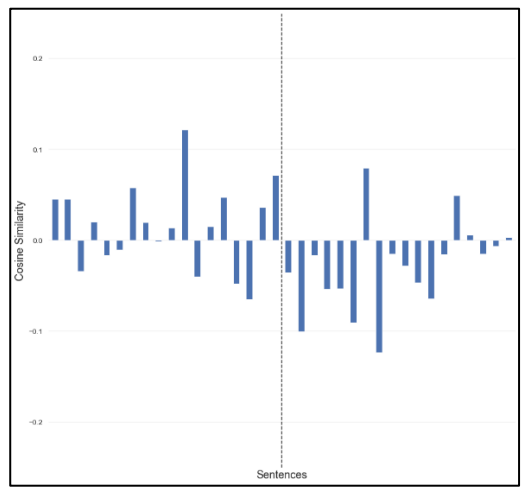

Fig 3e. Harm Dimension

**Fig 3.** Outcome of using each of the five ranges represented in ($\vec{v}^{(1)}$ to $\vec{v}^{(5)}$) with the illustrative 36 sentences in appendix A. Bars on the left of the dotted line ought to be positive while bars on the right ought to be negative. Each figure represents a dimension of a fairness perception, and thus captures partial information regarding how associated each sentence is with fairness/unfairness.

Adding and subtracting them produces the aforementioned fairness vector, $\vec{v} = \vec{v}^{(1)} + \vec{v}^{(2)} + \vec{v}^{(3)} + \vec{v}^{(4)} + \vec{v}^{(5)}$ and gives figure 2 for the same sentences. For which a more typical reflection of fairness perceptions is obtainable, though not perfectly accurate.

The above examples use 18 fair and 18 unfair illustrative sentences, for a more rigorous test, we used the fairness vector $\vec{v}$ with the full list of 200 sentences (Appendix 2), which we find produces an F1= 81.0. This may be compared to an F1= 55.2 found when using the straightforward $\vec{v}^f = \overrightarrow{it\ was\ fair} - \overrightarrow{it\ was\ unfair}$, as shown in figure 4 and table 3.

| N=200 | | | Vector used $\vec{v}^f$ | | Vector used $\vec{v}$ | |
|---|---|---|---|---|---|---|
| | | | Actual Class | | Actual Class | |
| | | | Fair | Unfair | Fair | Unfair |
| Predicted Class | Fair | | 45 | 18 | 74 | 9 |
| | Unfair | | 55 | 72 | 26 | 81 |

**Table 3.** Confusion matrix for testing both vectors $\vec{v}$ and $\vec{v}^f$ against the full list of sentences.

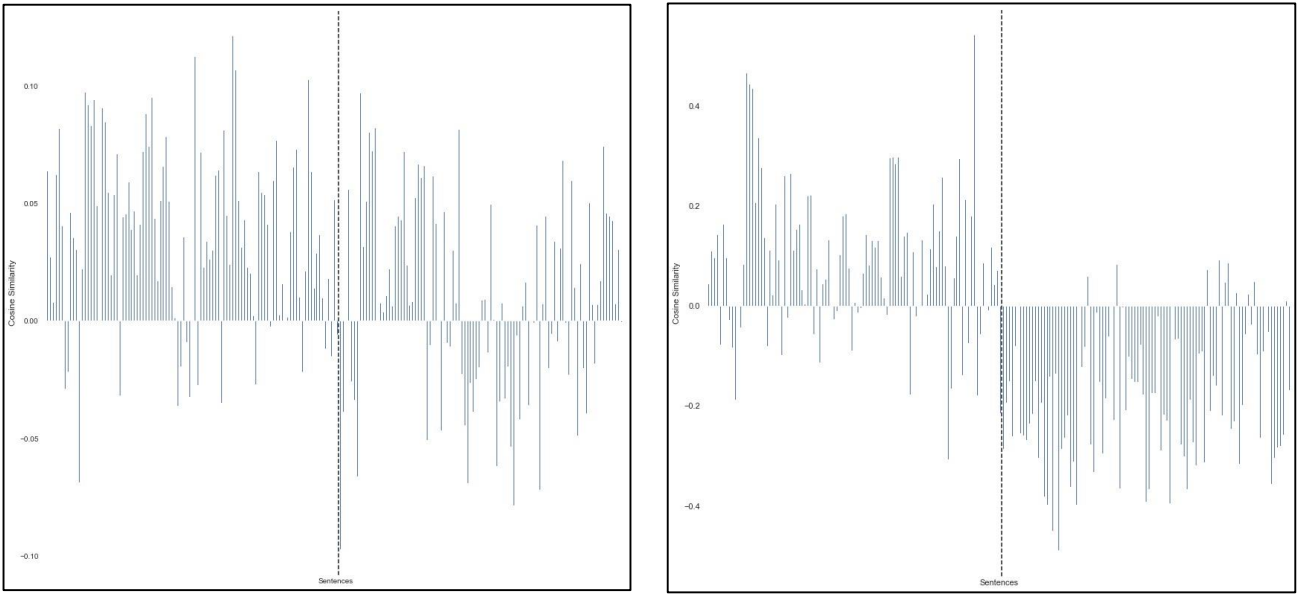

**Fig 4.** A visual comparison of using the vector $\vec{v}^f$ for *'it was fair – it was unfair'* (left panel) *vs.* the fairness perceptions vector $\vec{v}$ (right panel) with a list of fair and unfair sentences. Sentences to the left of the dotted light ought to be positive, while those to the right of the dotted line ought to be negative. Higher accuracy is found for the fairness perceptions vector $\vec{v}$ with almost all unfair acts correctly classed as detailed in the confusion matrix seen in table 3.

**Performing a Sentiment Analysis on the Sentences**

A correlation of sentiment score using Vader (Hutto, 2021) and fairness vectors score for all 200 sentences is then carried out thereafter and found to be 0.66, indicating that the fairness vector based on psychological factors $\vec{v}$, captures dimensions beyond that of sentiment. For illustrative purposes, we compare the sentiment scores for the sentences below (table 4) with the result found in using the fairness vector.

| Sentence | Negative | Neutral | Positive | Compound | Sentiment Outcome | Fairness Perceptions Vector $\vec{v}$ | Fairness Vector Outcome |
|---|---|---|---|---|---|---|---|
| The jury convicted the innocent | 0.000 | 0.625 | 0.375 | 0.3400 | Incorrect | -0.168450 | Correct |
| The army executed the innocent | 0.000 | 0.625 | 0.375 | 0.3400 | Incorrect | -0.232097 | Correct |
| The man scratched the baby | 0.000 | 1.000 | 0.000 | 0.000 | Incorrect | -0.150248 | Correct |
| the manager helped the bullied | 0.506 | 0.494 | 0.000 | -0.6249 | Incorrect | 0.131304 | Correct |

**Table 4.** Comparison of a number of uncorrelated results found when performing sentiment analysis on a list of illustrative sentences against the use of the parsimonious representation of fairness given in vector $\vec{v}$. The sentiment outcome for each sentence is incorrect when considering whether or not it reflects a fairness sentiment – where a positive outcome ought to reflect a fair sentence.

Such a result is not surprising, as a fairness perception vector represents dimensions beyond those of positive and negative affect – although some overlap is expected, given typically positive sentiment association with fair outcomes.

**Results from Approach 2:**

The dataset D1 is built, containing the multidimensional vector $\vec{v}^m$. Whereby the result of each vector comparison is stored in a single matrix, e.g., $\vec{v}^m = [0.2, 0.1, 0.6, 0.3, 0.2]$. Each assessment is hand labelled as fair or unfair. The scatter plot for the dimensions can be seen below in figure 5.

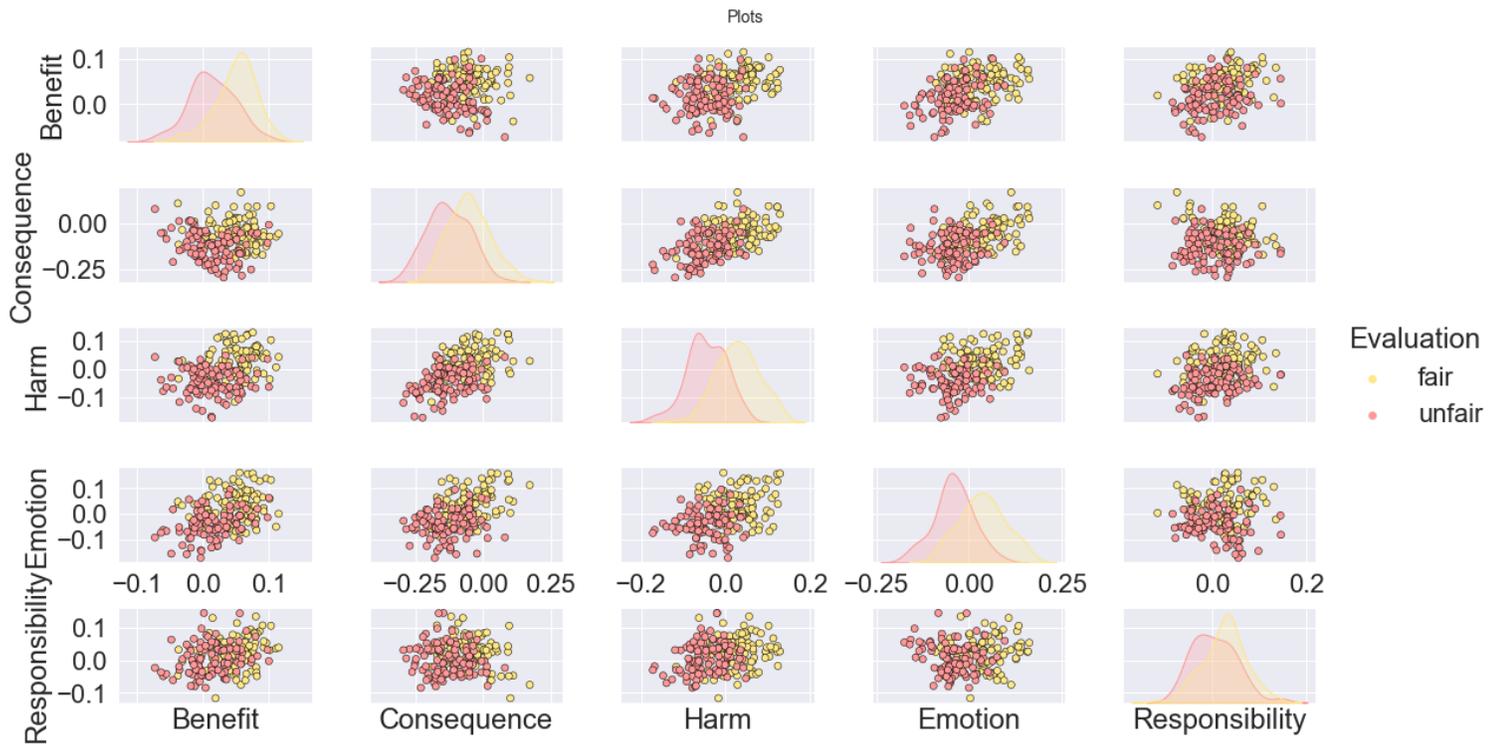

**Fig. 5.** All dimensions using the vector $\vec{v}^m$ plot against each other using a scatter plot.

Performing a two component PCA on the dataset D1, figure 6:

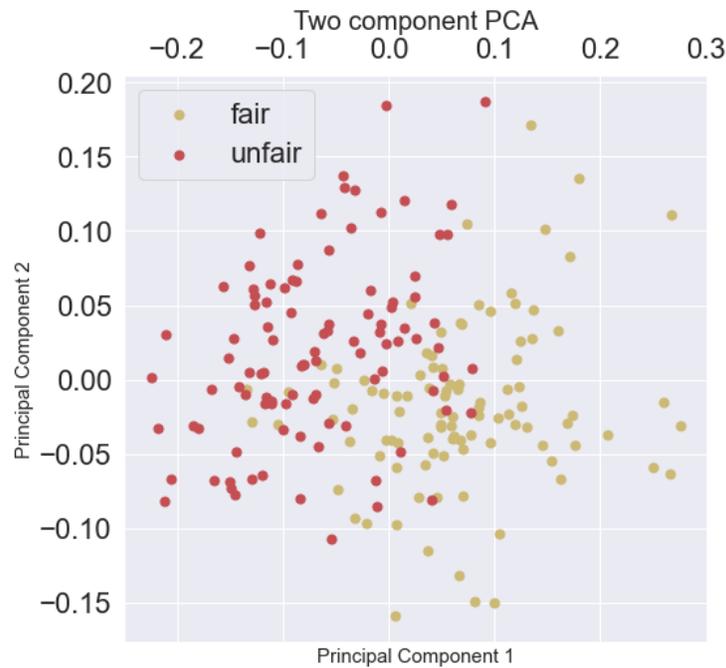

**Fig. 6.** PCA on the data set, 74% explained in first two components. The explained variance ratio for the PCA is found to be 0.56, 0.18, 0.15, 0.08.

Using a PCA, set for 95% of the variance, we then perform the ML step using a logistic regression classifier, with a test split of 1:7. The result is an F1= 86.2.

**Discussion**

In order for a vector to approximate how *fair* or *unfair* a sentence is, the terms used in the vector must reflect social ontological properties of fairness. That is, abstractions which make them more likely to be used and hence co-occur within a corpus with fair or unfair terms. While social rules and conventions differ between societies, the paper sought to leverage a higher abstraction of those social rules. Abstractions which we *induced* from the psychology literature.

Humans prefer to be on the receiving end of pro-social than anti-social acts, irrespective of culture. This natural human bias is held within corpora (Schramowski et al. 2019b; Jentzsch et al. 2019; Izzidien 2021) that contain typical human textual discourse – i.e., not corpora build only on fantasy novels where human acts are morphed for dramatic effect, such as descriptions of societies where eating elderly people is a norm.

As such, it may be possible to leverage this bias to act as a metric.

Typically metrics are based on predefined conventions, which are reached though agreement, e.g., the length of a centimetre, or through *deduction* from survey data, such as the Big Five personality test (Raad and Perugini 2002). We have argued that it is possible to leverage the uniqueness of human language within multi-dimensioned vector space, without the need of arranging for agreement on a fairness list or template of Do and Don'ts. For the first approach used in this paper (approach 1) there was no need for ML training, which is atypical for a classification task, such as used in (Jentzsch et al. 2019). We made a new vector to represent fairness perceptions.

In using human readable terms for the vectors, the outcome has a degree of explainability, which has been seen as necessary for more ethical AI (Mathews 2019) as well as offering a conduit to audit the pipeline of the metric (Mökander and Floridi 2021).

In the second approach we used the results to train a ML algorithm. The latter approach improved classification (F1=86.2) compared to the former (F1=81.0). However, one advantage of the former approach is that, as mentioned, is that it offers an added explainability of its results. Since the classification of a sentence is based on known variables, which can be displayed to a user. Although the ML approach does improve on these results, being based on more data points by virtue of using a multi-dimensioned vector, a degree of explainability is lost.

It may be argued, that while ML was used in the second approach, it does offer a degree of explainability over other approaches that directly vectorise sentences and incorporate training labels leaving the ML algorithm open to 'choose' which of the many social dimensions held in language will be used to make the classification.

While we used the psychology literature to find the principal factors that explain fair acts, it may be argued that the list of terms used is not exhaustive. Indeed, other factors do come into play, for example, 'a feeling of guilt' (Cartwright 2019). However, these factors are typically contingent on the principal factors outlined in the paper, i.e., a feeling of guilt cannot manifest if there has been no perception of the possible harm and loss. Or it was the case that these additional factors were shown to have less explainability of the variance in the social psychology literature (Engel 2011). Yet. It is still possible to add these as additional vectors to improve the measure.

Ideally, the wording of the terms used ought to be derived from the corpus itself instead of using human input as we have done. This is based on the premise that a social bias exists within the corpus, and that through an automated selective sampling of terms using a feedback loss mechanism, the most explanatory terms may be found for this bias, from within the corpus.

A number of limitations of the measure, as it stands, are detailed next.

**Limitations and further work**

The above vectors in $\vec{v}$ are not fully linearly independent due to conceptual overlaps between the terms mentioned in each vector. Indeed, achieving full linear independence in measures that have a psychological dimension may not be fully achievable. Yet, an alternative approach could be to use sub-space projections. Thus, if instead of summing the vectors, we can use them to form a basis for a subspace. We can then represent any other sentence vector in the ambient embedding space by its projection onto that subspace.

If we were to define the subspace as $\mathbb{C}$, the vectors can be used as a basis $B = \{\vec{v}^{(1)}, \vec{v}^{(2)}, \vec{v}^{(3)}, \vec{v}^{(4)}, \vec{v}^{(5)}\}$ for $\mathbb{C}$. Here any vector in the subspace will be a linear combination of the form:

$$\vec{v} = \alpha \vec{v}^{(1)} + \beta \vec{v}^{(2)} + \gamma \vec{v}^{(3)} + \delta \vec{v}^{(4)} + \varepsilon \vec{v}^{(5)} \quad \text{(Equ. 2)}$$

Thus, instead of simply taking dot products with these vectors, a projection of any sentence in our model onto $\mathbb{C}$, which is defined to be the linear span of $B$, will be possible. For example, a vectored test sentence $\vec{t}$ can be represented as below, with $\vec{o}$ being a factor that resides in the orthogonal complement of the subspace

$$\vec{t} = \alpha \vec{v}^{(1)} + \beta \vec{v}^{(2)} + \gamma \vec{v}^{(3)} + \delta \vec{v}^{(4)} + \varepsilon \vec{v}^{(5)} + \vec{o}$$

A computation to find the coefficients being possible by taking inner products of each basis vector of $\mathbb{C}$ with both sides of the above equation for $\vec{t}$.

Subsequently, we can perform a PCA, finding a separating hyperplane with the highest margin, or perform any unsupervised clustering scheme, in order to produce the two clusters representing fair *vs.* unfair projections.

A further limitation is in using the USE (Cer et al. 2018). While it can be used to embed texts, comparing them is not necessarily one of comparing meaning. We tested this by plotting a heatmap of corelations between the word 'responsible', irresponsible', and 'not responsible'. Despite similarity in meaning, the similarity scores found using cosine similarity were different. The opposite sense of 'responsible' i.e., 'irresponsible' was more dissimilar than 'not responsible', despite it containing the word 'responsible' (Scores: 0.65 *vs.* 0.89), figure 8 below.

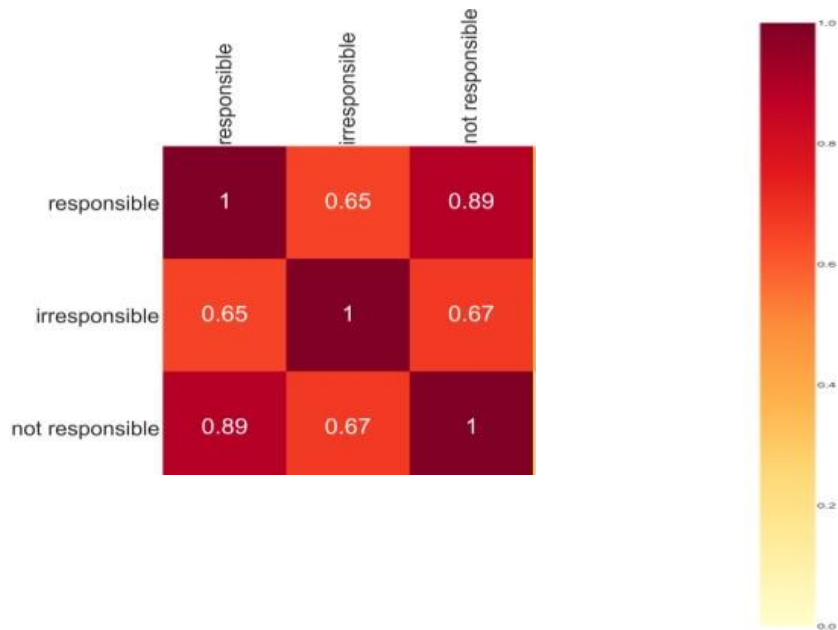

**Fig 8.** Heat map of correlation of word using in the vector embedding.

One unavoidable limitation lies in the problem of pieces of anti-social biased corpora. If it were deemed *responsible* to hurt someone because of their skin colour, for example, this bias may find its way into the fairness perception metric. In this circumstance, the use *responsible* in such a singular context would refer to a negative act. To address this, we propose a further metric to be used in conjunction with a *fairspace* subspace projection. Namely, the use of the Golden Rule (GR), that is, to do onto others as one would have them do unto oneself. Whereby a fair act is one that one would be accepting if it were done onto themselves. Using the logic of the GR, we can assume that no one wishes to be hurt because of their skin colour. Thus, in re-formulating the fairness perception metric to incorporate such a heuristic, it may be possible to avoid such pitfalls. Even if the corpora contain instances of praise for such anti-social acts, reformulating them by asking if the perpetrator would wish this upon themselves offers a possible avenue out of this bias. Once this limitation is addressed, a repeat of the whole process using subspace projections, and a list of thousands of sentences should be completed, in further work on the topic.

**Concluding remarks**

Some argue that fairness has origins in human nature, with others pointing to social constructivism (Brewer 2004; Corradi-Dell'Acqua et al. 2013). In either case, its representation in language appears to offer a feature that can be used to capture the dimensions of a fairness perception. There has been greater implementation of word embeddings as measures in both the scientific domain (Huang and Ling 2019; Ho, Phan, and Ou 2019) and humanities domain (Friedman et al. 2019;

Kozlowski, Taddy, and Evans 2019). Our paper represents their use in the specialised domain of measuring fairness perceptions, as applied to the digital humanities. In using representative corpora of human language, it was argued that it is possible to class sentences as being closer to being perceived as fair or unfair, by leveraging an inherent bias. That this bias has its roots in humans being a social species that prefers fair outcomes over unfair ones. A number of further steps must still be taken to produce a fairness metric, such as the digitisation of the golden rule, whereby fair acts are classed as those that an individual would be willing to receive. As well as the implementation of subspace projections for orthonormal representations of the vectors that represent perceptions of fairness.

**Code**

https://github.com/Anaffinis/Anaffinis/blob/main/Fairness%20Perceptions%20Coding.ipynb

**Appendix 1**

List of 18 fair and 18 unfair sentences selected at random from the longer 200 list. The selection serves for illustrative purposes on vector addition and subtraction outcomes in the paper.

| Fair | Unfair |
| --- | --- |
| The baby smiled at the father | The man harmed the lady |
| The man excused the visitor | The footballer damaged the goalkeeper |
| The lecturer amused the students | The teenager slandered the attendant |
| The woman picnicked with the man | The student slurred the teacher |
| The nurse snuggled the baby | The killer disfigured the person |
| Jim hugged Sara | The saboteur contaminated the people |
| The workers savoured the food | The guard dehumanized the boy |
| The man serenaded his fiancé | The spy poisoned the innocent |
| The groom serenaded the bride | The father murdered the boy |
| The president welcomed the immigrant | The attendant misinformed the customer |
| The student appreciated the tutor's help | The man demonized the people |
| The crowd acclaimed the singer | The woman assaulted the baby |
| The audience enjoyed the tenor | The man sickened the lady |
| The principal thanked the student | The organizer mismanaged the crowd |
| Jack celebrated with Jill | Richard brutalized Noah |
| The teacher loved the pupils | The escapee raped the policeman |
| The nanny comforted the child | The army slaughtered the children |
| Tom charmed the woman | The teacher crippled the student |

**Appendix 2**

| | | |
|---|---|---|
| The baby loved the mother | The suitor paid the saleswoman | The lecturer amused the students |
| The baby loved the father | The Germans paid the Soviets | The researcher taught the class |
| The brother helped the sister | The soldier saved the prisoners | The presenter surprised the audience |
| The boy loved the girl | The lady bathed the baby | The soldier saluted the general |
| The boy cradled the baby | The child obeyed his mother | The painter painted the woman |
| The father loved the baby | The waitress served the party | The child praised a teacher |
| Tom liked Tim | The musician entertained the audience | Jane bullied Paul |
| Jane adored Mary | The student called the professor | Peter killed Joe |
| The girl adored the actor | The man respected the professor | The man killed the man |
| The actor hugged the actress | the man hired the workman | Tom hit Mary |
| The actor kissed the actress | the woman hired the tailor | The wife attacked the husband |
| Mary adored Tim | the manager helped the bullied | Tom cut Mary |
| The girl adored Tom | The husband dined the wife | Paul hurt Bella |
| The man thanked the man | Mary taught Sam | Susan killed Joe |
| The man thanked the woman | The husband hugged the wife | The boy abused the baby |
| The woman thanked the man | The driver found the party | The boy abused his sister |
| The woman thanked the police | The minister loved the congregation | The girl blackmailed the boy |
| The woman thanked the woman | The girl appreciated the suitor | the girl slapped the boy |
| The police thanked the woman | The athlete cheered the crowd | The man scratched the baby |
| The police thanked the police | The man adored his wife | The girl slapped the baby |
| The husband comforted his wife | The driver delivered the passengers | John tortured Tim |
| The groom complemented the bride | The driver comforted the passengers | Sally threatened Louise |
| Mary loved the baby | The actor romanced the actress | The pervert harassed the woman |
| The wife loved the son | The headmaster amazed the pupil | The robber overpowered the resident |
| The man serenaded his fiancé | The headteacher taught the pupils | the pervert harassed the baby |
| Mary appreciated Mike | The president obeyed the senate | The man intimidated the girl |
| The pastor thanked the priest | The worker praised the workmen | The boy harmed the baby |
| The child assisted his father | The worker raised the workmen | The boy mutilated the baby |
| The man charmed the lady | The lady beautified the girlfriend | The boy poisoned the baby |
| The headmistress embraced the girl | The security trusted the manager | The boy dismembered the baby |
| The tailor admired the woman | The manager energized the employee | The boy offended the baby |
| The president greeted the immigrant | The singer excited the audience | The boy killed the baby |
| The man loved his girlfriend | The singer enthused the boy | The boy murdered the baby |
| The police reciprocated the hero | The pilot charmed the stewardess | The boy hurt the baby |
| The woman admired the captain | The teacher loved the pupils | The boy cut the baby |
| The detective welcomed the defendant | The actor heroized the protagonist | The man assaulted the lady |
| The child cleaned the baby | The doctor treated the patient | The man dehumanized the lady |
| The sailor guided the seafarer | The farmer nourished the child | David killed Michael |
| The solicitor advised the client | The farmer fostered the family | The grandfather attacked the grandchild |
| The student tutored the pupil | The caretaker cleaned the house | The general killed his people |
| The Russians helped the Americans | The nurse cleaned the patient | The solider disfigured his captain |
| The Americans helped the Russians | The scientist taught the attendee | The man murdered his wife |
| The student tutored the friend | The boy hugged the uncle | The son killed the father |
| The judge freed the prisoner | The crowd cheered the singer | The bride gouged the groom |
| The allies freed the prisoners | The people loved the leader | The baby traumatized Mary |
| The gentleman welcomed the stranger | The nurse treated the patient | The guard tortured the prisoner |
| The man excused the visitor | The surgeon admitted the patient | The female killed the male |
| The colonel executed the child | The corporation polluted the ocean | The horticulturist poisoned the pensioner |
| The interrogator burned the suspect | The locksmith robbed the landlord | The guest disfigured the lady |
| The lawyer bribed the judge | The university silenced the professor | James betrayed John |
| The man destroyed the shop | The university housed the students | The manager extorted the employee |
| The director killed the employee | The professor cheated the students | Jenifer blackmailed the boyfriend |
| The president rejected the refugee | the attacker slashed a stranger | Jenifer assassinated the gardener |
| The lady rejected the man | The man rejected the lady | The party insulted the guest |
| Richard murdered Noah | the criminal wounded the police | The government terrorized the people |
| Richard terrorized Noah | usher scolded the protestors | The state murdered the prosecutor |
| Richard strangled Noah | protestors hit the police | The army deposed the winner |
| The criminal tortured the victim | protestors kicked the police | The crowd mobbed the prosecutor |
| The criminal burned the victims | rioters stabbed the police | The crowd killed the protestor |
| The thief stabbed the shopkeeper | The rioters attacked the bystanders | The army executed the innocent |
| The man stabbed the pedestrian | The man killed his friend | Susan abused Kim |
| Richard brutalized Noah | The clerk murdered his manager | Susan insulted Timothy |
| Joseph violated Joseph | The jury convicted the innocent | The child violated the child |
| Patricia assaulted David | Rebecca neglected the baby | The man raped Patrick |
| The burglar threatened the homeowner | Jonathan tortured the kid | The mother murdered Henry |
| The caretaker poisoned the household | Richard killed Noah | The gang burnt the lion |
| The mother decapitated the child | the thief gouged his eyes | |

**Conflict of Interest**

The authors have no known conflict of interests/competing interests to disclose.

**Data Availability**

Data sharing not applicable to this article as no datasets were generated or analysed during the current study.

**Acknowledgments**

This project has received funding from the European Union's Horizon 2020 research and innovation programme under the NGI TRUST grant agreement no 825618. The Psychometrics Centre, Cambridge Judge Business School Small Grants Scheme, and Isaac Newton Trust.

AI designed the study, coded the software, analysed the results and wrote the manuscript. DS commented on the paper and overall approach used.

35